\documentclass[twoside,11pt]{article}

\usepackage[preprint,abbrvbib]{jmlr2e}
\usepackage{xcolor}
\usepackage{comment}
\usepackage{booktabs}
\usepackage{multirow}
\usepackage{listings}
\usepackage[frozencache=true,cachedir=.]{minted}
\ShortHeadings{gCastle: A Python Toolbox for Causal Discovery}{Zhang, Zhu, Kalander, Ng, Ye, Chen, and Pan}
\firstpageno{1}

\begin{document}

\title{gCastle: A Python Toolbox for Causal Discovery}

% More compact authors following: mvlearn, mlr3pipelines, pyDML
\author{\name Keli Zhang$\mathbf{^{1,*}}$ \email zhangkeli1@huawei.com \\
        \name Shengyu Zhu$\mathbf{^{1,*}}$ \email zhushengyu@huawei.com \\
        \name Marcus Kalander$\mathbf{^{1}}$ \email marcus.kalander@huawei.com \\
        \name Ignavier Ng$\mathbf{^{2}}$ \email ignavierng@gmail.com \\
        \name Junjian Ye$\mathbf{^{1}}$ \email yejunjian@huawei.com \\
        \name Zhitang Chen$\mathbf{^{1}}$ \email chenzhitang2@huawei.com \\
        \name Lujia Pan$\mathbf{^{1}}$ \email panlujia@huawei.com \\
       
        \addr \noindent $^1$ Noah's Ark Lab, Huawei Technologies, China\\
                        $^2$ University of Toronto, Canada\\
                        $^*$ Correspondence authors and equal contributions
       }

\editor{Kevin Murphy and Bernhard Sch{\"o}lkopf}

\maketitle

\begin{abstract}%   <- trailing '%' for backward compatibility of .sty file
\texttt{gCastle} is an end-to-end Python toolbox for causal structure learning. It provides functionalities of generating data from either simulator or real-world dataset, learning causal structure from the data, and evaluating the learned graph, together with useful practices such as prior knowledge insertion, preliminary neighborhood selection, and post-processing to remove false discoveries. Compared with related packages, \texttt{gCastle} includes many recently developed gradient-based causal discovery methods with optional GPU acceleration. \texttt{gCastle} brings convenience to researchers who may directly experiment with the code as well as practitioners with graphical user interference. Three real-world datasets in telecommunications are also provided in the current version. \texttt{gCastle} is  available under Apache License 2.0 at \url{https://github.com/huawei-noah/trustworthyAI/tree/master/gcastle}.
\end{abstract}


\section{Introduction}
Discovering and understanding causal relations underlying physical or artificial phenomena is an important goal in many empirical sciences. Due to the relative abundance of passively observed data as opposed to interventional data, learning causal graphs from purely observational data has been vigorously studied \citep{Peters2017book, spirtes1991algorithm}. Existing methods roughly fall into three classes: constraint-based, function-based, and score-based. The first class relies on conditional independence tests and identifies a class of Markov equivalent directed acyclic graphs (DAGs)~\citep{meek1995casual, spirtes2000causation, zhang2008completeness}. Unlike constraint-based methods that assume faithfulness and identify only the Markov equivalence class, function-based methods can distinguish between different DAGs in the same equivalence class, by imposing additional assumptions on data distributions and/or function classes. Examples include linear non-Gaussian additive model (LiNGAM)~\citep{Shimizu2006lingam,Shimizu2011directlingam} and the nonlinear additive noise model (ANM)~\citep{Hoyer2009NonlinearCausal}. The last class of methods employs a score function to evaluate candidate causal graphs w.r.t.~the data and then searches for a (or a class of) causal DAGs achieving the optimal score~\citep{Chickering2002optimal,Peters2017book}. Due to the combinatorial nature of the acyclicity constraint, most score-based methods use local heuristics to perform the search. Recently, a class of methods has considered differentiable score functions in combination with a novel smooth characterization of acyclicity~\citep{zheng2018dags}, so that gradient-based optimization method is feasible to seek the desired DAG. This change of perspective allows using deep learning techniques for flexible modeling of causal mechanisms and improved scalability. See, e.g., \citet{Yu19DAGGNN, Ng2019masked, Ng2019GAE, ng2020role,Lachapelle2019grandag, Zheng2019learning, Philippe2020DiffInterventional,Bhattacharya21DiffConfounder}, which have shown state-of-the-art performance in many experimental settings, with both linear and nonlinear causal mechanisms.

%functional causal models.
% Unlike constraint-based methods that assume faithfulness and identify only the Markov equivalence
% class, these methods are able to distinguish between different DAGs in the same equivalence class,
% thanks to the additional assumptions on data distribution and/or functional classes. Examples include
% LiNGAM (Shimizu et al., 2006; 2011), the nonlinear additive noise model (Hoyer et al., 2009; Peters
% et al., 2014; 2017), and the post-nonlinear causal model (Zhang and Hyvärinen, 2009)

This paper presents the {\bf gradient-based causal structure learning} (\texttt{gCastle}) toolbox, developed in Python and with PyTorch supporting GPU acceleration. \texttt{gCastle} aims to provide many ready-to-use gradient-based methods, but also classic and competitive algorithms such as PC~\citep{spirtes2000causation} and LiNGAM~\citep{Shimizu2006lingam,Shimizu2011directlingam}. The functionalities include dataset generation from either simulator or real-world datasets, causal structure learning, and evaluation of learned graphs. 
%\textcolor{blue}{marcus: the first part here is a bit repetition of what is in the related work section. Could be removed or adjusted to save space.}
 \texttt{gCastle} brings much convenience to researchers in the machine learning community who may directly experiment with the released Python code. For practitioners, \texttt{gCastle} provides useful practices in learning causal graphs, such as prior knowledge insertion, preliminary neighborhood selection to eliminate non-edges, and post-processing to remove false discoveries. A graphical user interference (GUI) is also developed to ease the causal structure learning process and to visualize the learned graph that allows further manual annotations.

In addition, \texttt{gCastle} releases three real-world causal datasets where the DAGs describe relations amongst alarms in telecommunication systems. The graphs are obtained by both human annotations and historical maintenance records (i.e., interventional data). We believe that these datasets would be a good addition to causal discovery research, as there are only a few public real-world datasets. The datasets were previously used in a causal discovery competition, with 112 participating teams and over 800 submissions in total.\footnote{\url{https://competition.huaweicloud.com/information/1000041487/introduction}}

\section{Related Packages}
There exist several Python packages focusing on causal inference, aiming at estimating the causal effect on the outcome of an intervention~\citep{causalml,dowhy,econml,whynot}. To the best of our knowledge, there are only four Python packages for causal structure learning: Tigramite~\citep{tigramite}, py-causal~\citep{py-causal}, causal-learn~\citep{causal-learn}, and CDT~\citep{Kalainathan2020CDT}. Tigramite has a narrow focus on time series data and is not directly comparable to \texttt{gCastle}. py-causal is a wrapper around the Java-based Tetrad~\citep{tetrad} package, and all the algorithms and relevant functionality are called from Java, resulting in somewhat inconvenience for use and further modifications. Similarly, CDT is not a pure Python package and has a significant part of its algorithms and evaluation metrics called from R code.
On the other hand, causal-learn is a recently released direct Python translation of the Tetrad code, however, no gradient-based methods are included.

A slight overlap exists between the algorithms implemented in \texttt{gCastle} and those offered by CDT, causal-learn, and py-causal. The overlap is mainly present in the traditional methods that are often used as baselines in this field. \texttt{gCastle} has a clear focus on more recently developed gradient-based structure learning algorithms which are missing from the other packages.

\section{Design and Implementation}
\begin{figure}[t]
    \centering
    \includegraphics[width=.9\textwidth, trim={0 0 0 0}, clip]{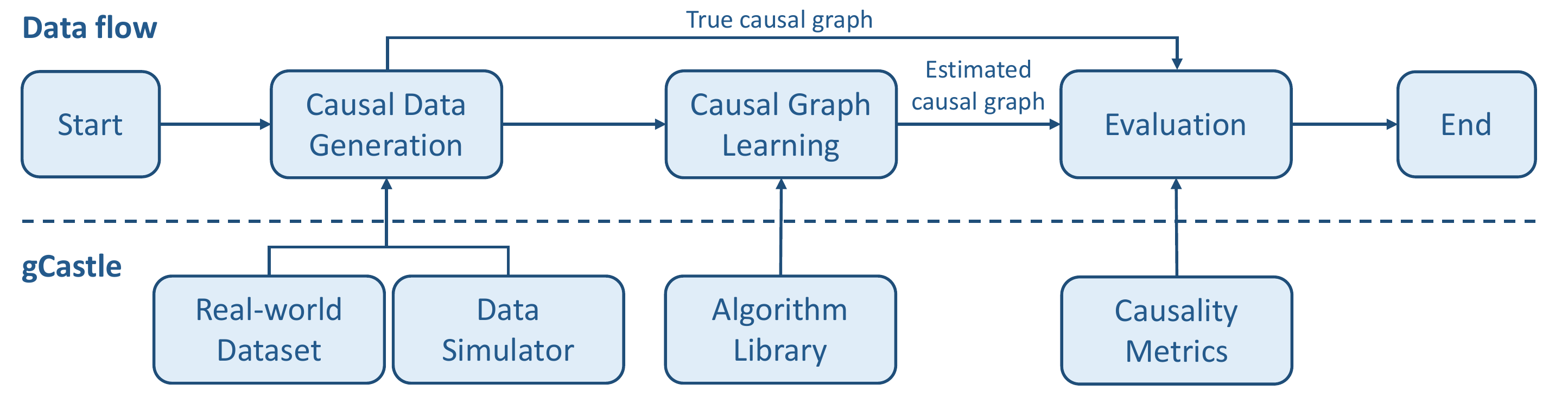}
    \caption{Causal discovery workflow.}
    \label{fig:workflow}
\end{figure}

%\subsection{Workflow}
%% maybe remove this section title?
The vision of \texttt{gCastle} is to provide an end-to-end pipeline to ease causal discovery tasks that allow: simulating causal data or loading real-world data; learning causal graphs with state-of-the-art algorithms like recent gradient-based algorithms; evaluating estimated causal graphs with commonly used metrics such as false discovery rate (FDR), true positive rate (TPR) and structural Hamming distance (SHD); and using a user-friendly web interface to visualize the whole procedure. Figure~\ref{fig:workflow} illustrates the overall workflow.

\begin{figure}[t]
    \centering
    %  trim={<left> <lower> <right> <upper>}
    \includegraphics[width=0.85\textwidth, trim={35pt 25pt 20pt 28pt}, clip]{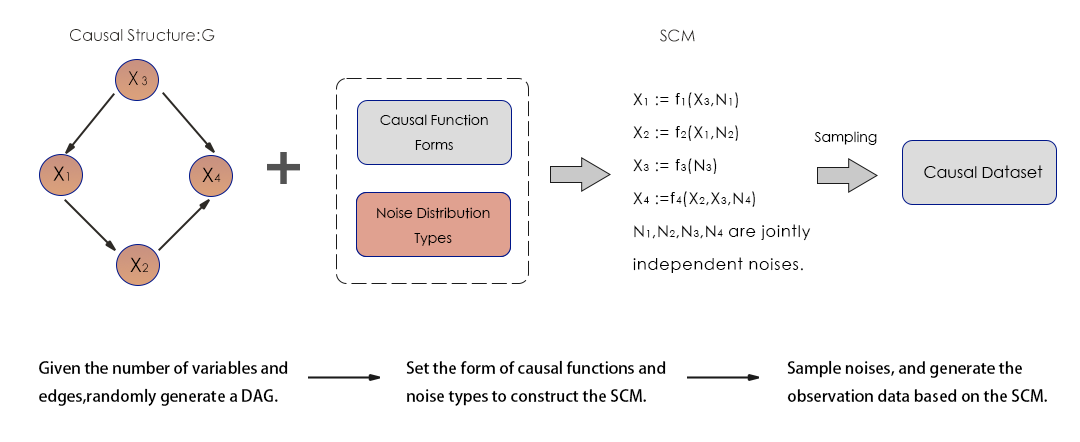}
    \caption{Synthetic data simulation procedure.}
    \label{fig:simulation_procedures}
\end{figure}

\subsection{Datasets}
\texttt{gCastle} provides various types of data generators to help users simulate data or load real-world data. % In the data generator implementation, we formalize the data simulation procedure as illustrated in Figure~\ref{fig:simulation_procedures}.
Currently, three real-world datasets are provided and each contains observational records collected from devices in real telecommunication networks and a true causal graph labeled by business experts. The datasets can be accessed through the \texttt{gCastle} API.
For simulated data, the procedure is illustrated in Figure~\ref{fig:simulation_procedures}.
Provided with a random DAG, causal function forms, noise distribution, and predefined sample size, one can quickly simulate a synthetic observational dataset that follows the given structural causal model (SCM). At present, \texttt{gCastle} supports multiple well-known causal functions, including Linear, MLP, and Quadratic, and also various noise distribution types, such as Gaussian, Exponential, Uniform, and Gumbel. Additionally, \texttt{gCastle} offers functionality to generate random DAGs with different strategies, which include ER, scare-free, low-rank, etc.
%Currently, \texttt{gCastle} also provides three real-world datasets, each of them includes a fair number of observation records collected from devices in a real telecommunication network and a true causal graph labeled by business experts, and users can access these datasets in a standard format using the gCastle API.

%\textcolor{blue}{shengyu: shall we remove this section? considering the limited page and also we have mentioned in the intro}
%Besides, considering the gap between theoretical causal assumptions and real-world causal mechanisms, and to better instruct researchers to design more rational algorithms, \texttt{gCastle} also provide three real-world datasets, each of them includes a fair number of observation records collected from devices in a real telecommunication network and a true causal graph labeled by business experts. 
    
\begin{table}[t]
\footnotesize
\centering
\caption{List of algorithms in \texttt{gCastle}.}
\label{tab:alg_list}
\begin{tabular}{c|p{11.9cm}}
\toprule
\textbf{Category} & \multicolumn{1}{c}{\textbf{Algorithms}} \\ \midrule
\multirow{2}{*}{constraint-based} & original-PC~\citep{kalisch2007estimating}, stable-PC~\citep{colombo2014order}, parallel-PC~\citep{le2016fast} \\
\midrule
\multirow{2}{*}{function-based} & Direct-LiNGAM~\citep{Shimizu2011directlingam}, ICA-LiNGAM~\citep{Shimizu2006lingam}, ANM~\citep{Hoyer2009NonlinearCausal}, HPCI~\citep{zhang2020influence} \\ 
\midrule
\multirow{1}{*}{score-based} & GES~\citep{Chickering2002optimal}, TTPM~\citep{cai2021thp} \\
\midrule
%gradient-based & \begin{tabular}[c]{@{}l@{}}GraN-DAG~\citep{lachapelle2019gradient}, NOTEARS~\citep{zheng2018dags}, NOTEARS-MLP~\citep{Zheng2019learning}, NOTEARS-SOB~\citep{Zheng2019learning},\\ NOTEARS-LOW-RANK~\citep{fang2020low}, NOTEARS-GOLEM~\citep{ng2020role}, MCSL~\citep{Ng2019masked}, GAE~\citep{Ng2019GAE},\\ RL~\citep{zhu2019causal}, CORL~\citep{wang2021ordering}
% \end{tabular} \\

\multirow{4}{*}{gradient-based} & GraN-DAG~\citep{Lachapelle2019grandag}, NOTEARS~\citep{zheng2018dags}, NOTEARS-MLP~\citep{Zheng2019learning}, NOTEARS-SOB~\citep{Zheng2019learning}, NOTEARS-LOW-RANK~\citep{fang2020low}, NOTEARS-GOLEM~\citep{ng2020role}, MCSL~\citep{Ng2019masked}, GAE~\citep{Ng2019GAE}, RL-BIC~\citep{Zhu2020causal}, CORL~\citep{wang2021ordering} \\
\bottomrule
\end{tabular}
\end{table}
\subsection{Algorithms}
% https://english.stackexchange.com/questions/113422/how-to-use-hyphens-appropriately-when-listing-multiple-hyphenated-terms
So far, \texttt{gCastle} (version 1.0.3) implements 19 causal discovery algorithms which cover most gradient-based algorithms and some traditional constraint-, score-, and function-based algorithms. 
Compared to other mainstream toolboxes, \texttt{gCastle} has a fairly complete gradient-based algorithm library.
Table~\ref{tab:alg_list} lists all algorithms that \texttt{gCastle} currently supports.

\subsection{Evaluation}
The current version of \texttt{gCastle} provides nine metrics for evaluating the estimated causal graph relative to the underlying truth. Most metrics are commonly used in the literature, e.g., FDR, TPR, SHD, etc. Other metrics are designed according to the specific purpose in real-world scenarios; a relevant example is the \textit{gScore} which comes from a root cause analysis scenario in AIOps. 

\section{Installation and Usage}
\texttt{gCastle} can be installed locally using either \texttt{pip} or running the \texttt{setup.py} script in the source code. After installation, the gCastle API can be used to build tasks; an example using Notears is shown in Code Listing~\ref{listing:api_example}. An alternative to the main API is to use the provided user-friendly GUI to visually design the tasks\footnote{A demo is available at \url{https://www.youtube.com/watch?v=5NOu2oApBgw}.}. A Docker image with the GUI is available on Docker Hub\footnote{\url{https://hub.docker.com/r/gcastle/castleboard-cpu-torch}}, which can be used to avoid issues with software dependencies and version matching.

\begin{listing}[t]
\begin{minted}[fontsize=\footnotesize]{python}
from castle.common import GraphDAG
from castle.metrics import MetricsDAG
from castle.datasets import IIDSimulation, DAG
from castle.algorithms import Notears

# I. Generate the artificial true causal graph and observation data.
weighted_random_dag = DAG.erdos_renyi(n_nodes=10, n_edges=20, 
                                      weight_range=(0.5, 2.0), seed=1)
dataset = IIDSimulation(W=weighted_random_dag, n=2000, method='linear', 
                        sem_type='gauss')
true_causal_matrix, X = dataset.B, dataset.X
# II. Learn the Causal Structure beneath the observation data. 
nt = Notears()
nt.learn(X)
# III. Visualize the comparison of estimated/true graphs using a heat map.
GraphDAG(nt.causal_matrix, true_causal_matrix)
# IV. Calculate Metrics.
mt = MetricsDAG(nt.causal_matrix, true_causal_matrix)
print(mt.metrics)
\end{minted}
\caption{A toy example using the gCastle API.}
\label{listing:api_example}
\end{listing}
%hold
% \textcolor{red}{shengyu: may be something like below?} \textcolor{blue}{The pyDML library can be installed through PyPI (Python package index ), using the command pip install pyDML. It is also possible to download or clone the repository directly
% from GitHub. In such a case, the installation of the software package can be done by running the setup script available in the root directory, using the command python setup.py
% install. Once installed, we can access all DML algorithms, and the additional functionalities, by importing the desired class within the dml module}

\section{Concluding Remarks  and Future Developments}
This paper introduces a causal structure learning toolbox \texttt{gCastle}. It provides many recently developed gradient-based causal discovery methods and all the algorithms are implemented in Python. As an open-source software, \texttt{gCastle} encourages contributions of algorithms and datasets from both research and industry communities. Our future work direction is to continuously include more real datasets as well as add  competitive algorithms like CAM~\citep{Buhlmann2014cam} which may require re-implementation using Python.

%that is developed in 
%The \texttt{gCastle} software has been used by many and the datasets were also used in a competition. We have received many emails with inquires to use the released datasets in ongoing projects and research. Our future plan is to continuously include more recent and innovative algorithms but also more real datasets.

% The sections after the introduction are quite diverse among accepted ML open-source papers. Which ones to include depends a bit on what we want to highlight. Some possible suggestions (no particular order):
% Design / API design / architecture.
% Software features (or library overview / software description). What features or algorithms are supported. 
% Comparison with others (e.g., performance / running times / resource usage or algorithm / feature availability. Either through a table or experimental results.
% Related work / Comparison to related software. Similar software and how we differ. 

% Some papers have these as well (less interesting from a reader perspective):
% Installation and usage
% Quality standards / Project development / project assets. Code coverage, Travis CI, PEP8, documentation, etc.
% Dependencies / resources

% Extreme examples:
% scikit-network: https://jmlr.csail.mit.edu/papers/volume21/20-412/20-412.pdf
% chainerRL: https://jmlr.csail.mit.edu/papers/volume22/20-376/20-376.pdf

% Possible reference that we could follow:
% mvlearn: https://jmlr.csail.mit.edu/papers/volume22/20-1370/20-1370.pdf

% \vskip 0.2in
\def\bibfont{\small}
\bibliography{refs, szhu, zkl}

\end{document}